\documentclass[conference,letterpaper]{IEEEtran}
\IEEEoverridecommandlockouts
\usepackage[dvips]{graphicx} 
\usepackage{multirow}
\usepackage[left=0.71in,top=0.94in,right=0.71in,bottom=1.18in]{geometry}
\usepackage{times} 
\usepackage{amsmath} 
\usepackage{amssymb}  
\usepackage{subcaption}
\usepackage{epsfig}
\usepackage{float}
\usepackage{caption}
\usepackage{graphicx}
\usepackage{multirow}
\usepackage{bm}
\usepackage{booktabs}
\usepackage{subcaption}
\usepackage{xcolor}
\usepackage{color}

\begin{document}

\title{\huge Efficient Robotic Task Generalization Using Deep Model Fusion Reinforcement Learning}

\author{Tianying Wang$^{1}$, Hao Zhang$^{1}$, Wei Qi Toh$^{1}$, Hongyuan Zhu$^{1,3}$,\\ Cheston Tan$^{1,3}$, Yan Wu$^{1,3}$, Yong Liu$^{1,2}$,  Wei Jing$^{1,3,*}$ 
\thanks{$^{1}$ A*STAR Artificial Intelligence Initiative (A*AI); 1 Fusionopolis Way, 138632, Singapore. }
\thanks{$^{2}$ Institute of High Performance Computing; 1 Fusionopolis Way, 138632, Singapore. }
\thanks{$^{3}$ Institute of Information Research; 1 Fusionopolis Way, 138632, Singapore. }
\thanks{$^{*}$ Email address: {\tt\small 21wjing@gmail.com}}
}

\maketitle
\begin{abstract}
Learning-based methods have been used to program robotic tasks in recent years. However, extensive training is usually required not only for the initial task learning but also for generalizing the learned model to the same task but in different environments. In this paper, we propose a novel Deep Reinforcement Learning algorithm for efficient task generalization and environment adaptation in the robotic task learning problem. The proposed method is able to efficiently generalize the previously learned task by model fusion to solve the environment adaptation problem. The proposed Deep Model Fusion (DMF) method reuses and combines the previously trained model to improve the learning efficiency and results. Besides, we also introduce a Multi-objective Guided Reward (MGR) shaping technique to further improve training efficiency. The proposed method was benchmarked with previous methods in various environments to validate its effectiveness.
\end{abstract}


\section{Introduction}

As a result of the rapid development of robotic technologies, robots have been widely used in various applications in recent years. Nevertheless, programming the robot for given tasks is still manual and costly. For many robotic applications, extensive manual teaching or offline programming is required to program robotic tasks. Even for the same category of robotic tasks with a slightly different environment, trajectory adjusting or fine-tuning is still required. To improve efficiency and reduce the effort required for robotic programming, learning-based methods can be used to program the robotic tasks. With learning-based methods, robot programming can be done without much manual programming or tuning of the trajectories. Compared to traditional manual teaching or offline programming, learning-based methods are also more general and robust to handle the same category of tasks. 

Among the learning-based methods, Reinforcement Learning (RL) is one commonly used method to tackle the robotic task learning problem \cite{kober2013reinforcement, polydoros2017survey}. RL algorithms have been successfully applied to robotic applications such as assembly~\cite{fan2019learning}, pouring~\cite{do2018learning}, and insertion~\cite{lee2019making} in recent years. The robots first learn a policy with RL, then with the learned policy, they can generate actions based on the current state/observations. The RL-based methods usually work well for the same tasks with similar environment settings after extensive initial training. 

However, in many robotic programming applications, the environment may change over time for the same robotic task (e.g. pushing, grasping). The performance of the learned policy may thus drop for the same robotic task, when applying the learned model to the changed environment \cite{lee2019making,fan2019learning}. Therefore, additional training would be required for the learned model to generalize and adapt to the new environment in order to improve the results. Such an additional training for task generalization and environment adaptation is usually costly. Thus, reducing the cost of additional training for task generalization and environment adaptation is desired, to improve efficiency.
\begin{center}
\begin{figure}[thpb]
    \centering
    \includegraphics[trim={0cm 0cm 0cm 0cm},clip,width=0.49\textwidth]{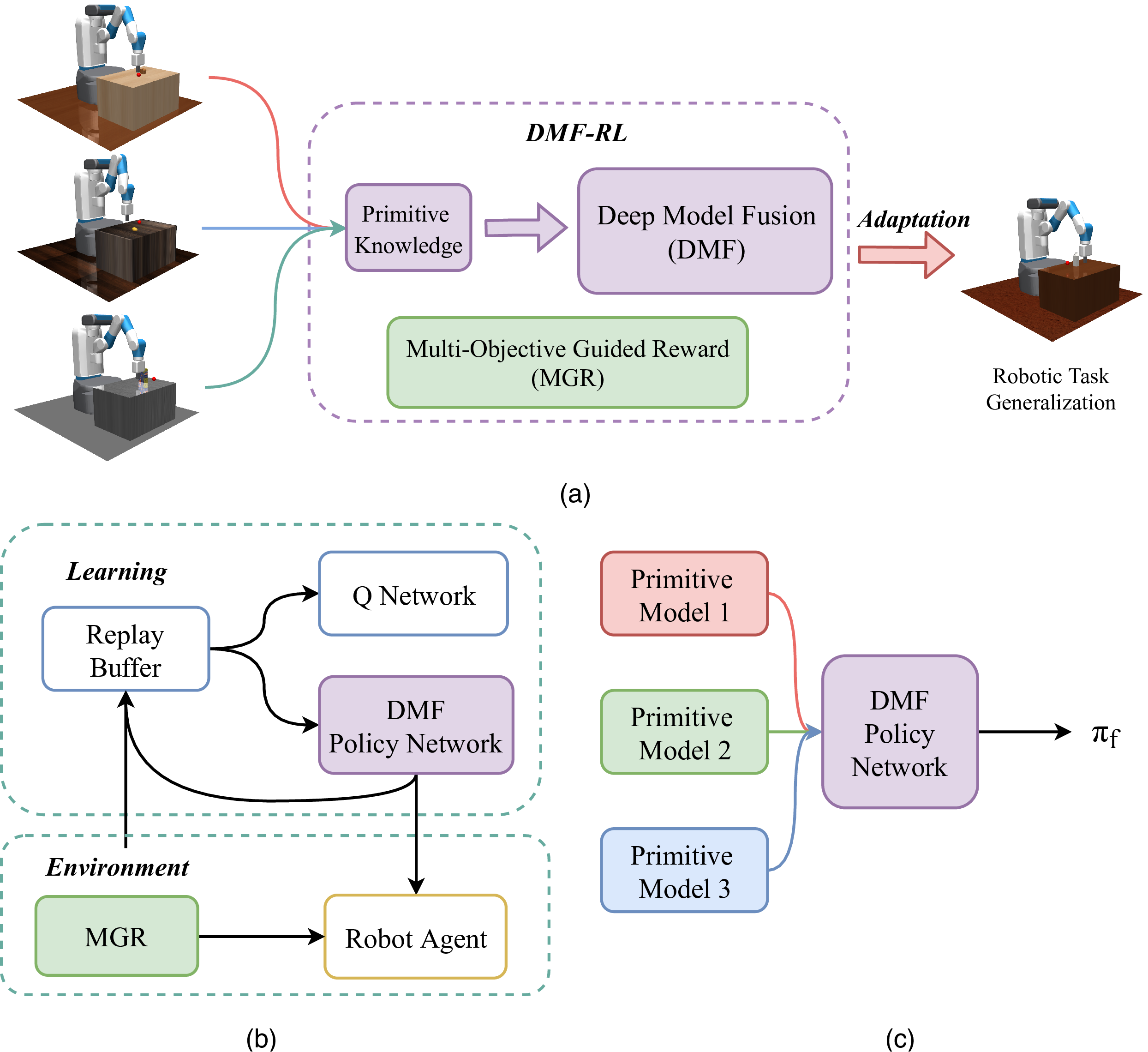}
    \caption{
    (a) The DMF-RL system that combines the knowledge from previous learned model for efficient robotic task generalization (b) The robot agent learning structure using DMF-RL system (c) DMF policy network structure
    }
    \label{fig_abstract}
\end{figure}
\end{center}
In this paper, we propose a novel algorithm for efficiently generalizing the robotic task learning problem to a new environment. The proposed method uses model fusion to reuse the previously learned knowledge and thus speed up the training in the task generalization process and improve the system performance. The main contributions of the proposed method are: 
\begin{itemize}
    \item a Deep Model Fusion (DMF) method to store and combine the previous trained knowledge, which speeds up the training process for task generalization and improves the results when the environment changes;
    \item a Multi-objective, Guided Rewards (MGR) system that converts the sparse rewards of typical RL problem to a multi-objective dense rewards system; and
    \item extensive studies to benchmark and validate the proposed methods in different environment settings. 
\end{itemize}

\begin{center}
\begin{figure*}[thpb]
    \centering
    \includegraphics[trim={0cm -1cm 0cm 0cm},clip,width=1\textwidth]{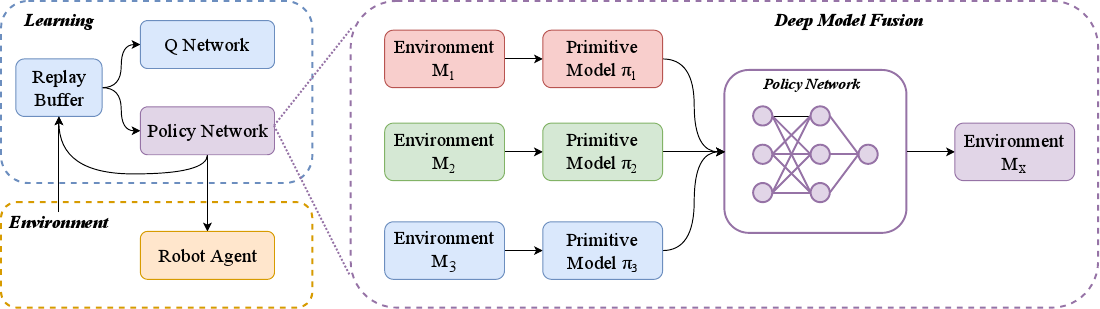}
    \caption{
    Deep Model Fusion Reinforcement Learning Architecture.
    }
    \label{fig_architecture}
\end{figure*}
\end{center}

\section{Related Work}

Reinforcement Learning (RL) has demonstrated great success in many applications during the past few years ~\cite{kober2013reinforcement, mnih2015human, silver2016mastering}. Value-based RL methods such as Deep Q-Learning (DQN) and its variants have outperformed human beings in various Atari Games~\cite{mnih2015human} \cite{van2016deep}. Value-based RL methods employ a $argmax$ to select the action with the maximum Q-value, which makes the value-based methods more suitable for the applications with discrete action spaces. RL has also been applied to many robotic learning problems \cite{chatzilygeroudis2018survey, ding2018hierarchical} with continuous action spaces. Because of the continuous action space in robotic applications, policy-based methods are more suitable and are more commonly used \cite{sutton2000policy}. The policy-based methods compute the gradient of the parameterized policy and improve the policy using the policy iteration mechanism \cite{sutton1998introduction}. Policy-based algorithms such as Deterministic Policy Gradient~\cite{silver2014deterministic}, Deep Deterministic Policy Gradient (DDPG) \cite{lillicrap2015continuous} and Proximal Policy Optimization (PPO) \cite{schulman2017proximal} have been successfully applied to address the continuous action space problems in various robotic applications. 

RL usually requires a large amount of data to train the model. In order to improve the training efficiency, different importance sampling methods have been proposed to sample from the experience into the replay buffer, such as prioritized replay buffer \cite{schaul2015prioritized} and Hindsight Experience Replay (HER)~\cite{andrychowicz2017hindsight}. Asynchronous Advanced Actor-Critics (A3C) \cite{mnih2016asynchronous} has also been proposed to parallelize and improve the computational efficiency of training. Generalization and adapting to a new environment is also an important research direction for RL applications in robotics \cite{hester2010generalized, finn2016generalizing}. The performance of the previously trained model would drop even when applying it to similar tasks with different environment settings. Methods such as training from scratch, or additional training from the existing model, could be used to train the model to improve the performance. Transfer learning could also be applied to RL problems as well \cite{cheng2017autonomous}. However, these learning paradigms are not designed to handle the environment adaption problems in robotic task learning applications; and they still require extensive training when the environment changes. In this paper, we propose a model fusion method to reuse the previously trained knowledge to improve the training efficiency and system performance when the environment changes. 

\section{Proposed Method}

In this paper, we propose a novel Deep Model Fusion Reinforcement Learning (DMF-RL) method for efficient robotic task generalization. The proposed method aims to improve efficiency when generalizing the learned task to similar tasks with different environments settings. The proposed DMF method combines knowledge from the previously trained models to reduce the training required for task generalization. A multi-objective guided rewards system is also proposed alongside the method to convert the sparse rewards to dense rewards and thus further speeds up the training process. The proposed method is illustrated in Figure~\ref{fig_architecture}.

\subsection{Markov Decision Process}

A finite-horizon Markov Decision Process (MDP) is used to model the robot task learning problem in this paper. MDP could be represented as a tuple $(\mathcal{S}, \mathcal{A}, \mathcal{T}, r, \lambda)$, where $\mathcal{S} $ is the state space; $\mathcal{A} $ is the action space; $\mathcal{T} : \mathcal{S} \times \mathcal{A} \Rightarrow \mathcal{S}$ is the state transition model; $ r: \mathcal{S} \times \mathcal{A} \Rightarrow r \in \mathbb{R}$ is the rewards by taking an action at a certain state; and $\lambda \in [0,1]$ is the discount factor. The return ${R}=\sum_{i=0}^N \lambda^i r_i$ of an episode is the summation of discounted rewards received during the episode. 

For the robotic task learning applications discussed in this paper, the state space $s \in \mathcal{S}$ is a 1-D vector that consists of the robot joint angles and joint velocities, as well as the current positions, orientations, and velocities of the objects. 

\subsection{RL for Robotic Task Learning}

With the MDP formulation, RL algorithms could be used to train the agent for the robotic task learning. As shown in Figure~\ref{fig_architecture}, we utilized an actor-critic RL framework, where a Q-network (critic) is used to approximate the Q-value, and a policy network (actor) is used to generate the action based on the current state. Q-learning is adopted to update the Q-value, and the policy gradient is computed to update the policy network. 

\subsection{Policy Network with Deep Model Fusion}

In this paper, Deep Model fusion (DMF) is proposed to reuse previously trained knowledge in the policy network, in order to improve the training efficiency and improve the model performance. With the proposed DMF, we use primitive policy models learned from several different environments in previous training. The fusion model is embedded as the policy network to be trained on the robot agent in the changed environment. With the primitive knowledge embedded in the fusion model, the agent robot is capable of adapting to new environments rapidly with better performance. 

Typically, the primitive knowledge is generated by training the robot agents in several  environments with different features. Suppose those environments, $\{\mathcal{M}_{p_1}, \mathcal{M}_{p_2}, \dots, \mathcal{M}_{p_n}\}$, are identical except the state transition probabilities $\mathcal{P}(s_{t+1}|s_{t},a_{t})$.

For the actor-critic RL algorithm such as DDPG, the policy is represented as $\mathcal{\pi}$ and the policy network is usually a neural network with parameters $\mathcal{\theta}_{\pi}$. By training the robot agent under different environments $\{\mathcal{M}_{p_1}, \mathcal{M}_{p_2}, \dots, \mathcal{M}_{p_n}\}$, we can obtain different  policy models whose policies and network parameters are denoted as $\{\mathcal\pi_1, \mathcal\pi_2, \dots, \mathcal\pi_n\}$ and $\{\mathcal{\theta}_{\pi_1}, \mathcal{\theta}_{\pi_2}, \dots, \mathcal{\theta}_{\pi_n}\}$, respectively.

\begin{center}
\begin{figure}[thpb]
    \centering
    \includegraphics[trim={0cm 0cm 0cm 0cm},clip,width=0.49\textwidth]{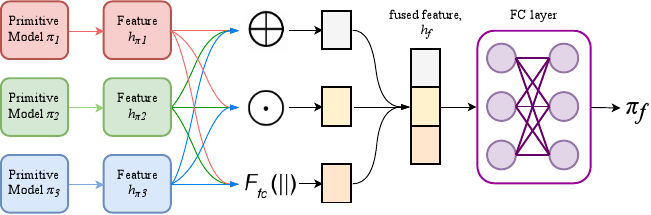}
    \caption{The Architecture of Policy Network with Deep Model Fusion.}
    \label{fig_method}
\end{figure}
\end{center}

When the environment changes, the performance of the learned policy may drop. The new policy model $\mathcal\pi_f$ in our DMF-RL method is then generated to improve the performance by fusion of the  policy models $\{\mathcal\pi_1, \mathcal\pi_2, \dots, \mathcal\pi_n\}$, as shown in Figure~\ref{fig_method}. 

Taking a three-model fusion case as an example, the model $\mathcal\pi_f$ first loads the parameters of $1st$ layer from each  policy model as $\mathcal{\theta}'_{\pi_1}$, $\mathcal{\theta}'_{\pi_2}$, and $\mathcal{\theta}'_{\pi_3}$, respectively. Since the $1st$ layer of the  policy networks encode low-level features from observations and the observation spaces in  models are identical, we extract those $1st$ layer features, denoted as $\{h_{\pi_1}, h_{\pi_2}, h_{\pi_3}\}$, from the $\{\mathcal{\theta}'_{\pi_1}, \mathcal{\theta}'_{\pi_2}, \mathcal{\theta}'_{\pi_3}\}$ and treat them as the primitive knowledge of the environments. All of the environmental features of  models contain useful information, so we combine them to further boost the performance of $\mathcal\pi_f$.

Given the $1st$ layer features, $\{h_{\pi_1}, h_{\pi_2}, h_{\pi_3}\}$, extracted from the previously trained  models, we use element-wise addition ($\oplus$), element-wise multiplication ($\odot$) and concatenation ($\parallel$) followed by a fully connected layer ($\mathcal{F}_{fc}$) to fuse the feature information. Typically, addition and multiplication operations allow additive and multiplicative interaction among different features without changing the feature dimension $d$. Although concatenation operation will double the feature dimension, the $\mathcal{F}_{fc}$ allows interaction among all elements and then maps to the original feature dimension $d$. After the three operations, we directly concatenate the generated features as the input of the new policy model $\mathcal\pi_{f}$. In general, the policy $\mathcal\pi_f$ after mode fusion is formulated as:

\begin{equation}
\begin{split}
    h_f = &(h_{\pi_1} \oplus h_{\pi_2} \oplus h_{\pi_3}) \\
    & \mathbin\Vert (h_{\pi_1} \odot  h_{\pi_2} \odot h_{\pi_3}) \\ &  \mathbin\Vert \mathcal{F}_{fc}(h_{\pi_1} \mathbin\Vert h_{\pi_2} \mathbin\Vert  h_{\pi_3}).
\end{split}
\end{equation}
where $\mathcal{F}_{fc}(x)=\boldsymbol{\omega}^{\top}_{fc}x + \mathbf{b}_{fc}$, $\boldsymbol{\omega}_{fc}\in\mathbb{R}^{3d\times d}$ and $\mathbf{b}_{fc}\in\mathbb{R}^{d}$ are the weight and bias of full connected layer, respectively.

For more general case, $\mathcal{N}$  models are represented in a sequence $\mathbf{\Theta} = (\theta'_{\pi_1}, \dots, \theta'_{\pi_n})$ and the corresponding features are denoted as $\mathbf{h}=(h_{\pi_1}, \dots, h_{\pi_n})$, then the fused feature of $\mathcal\pi_f$ is represented as:

\begin{equation}
\begin{split}
    h_f = & (h_{\pi_1} \oplus \dots \oplus h_{\pi_n}) \\ 
    & \mathbin\Vert (h_{\pi_1} \odot  \dots \odot h_{\pi_n}) \\ 
    & \mathbin\Vert \mathcal{F}_{fc}(h_{\pi_1} \mathbin\Vert \dots \mathbin\Vert  h_{\pi_n}).
\end{split}
\end{equation}

Finally, the fused feature $h_f$ is fed into a fully connected layer of a neural network to build up the fusion model policy $\pi_f$. The DMF-RL framework takes $\pi_f$ as the policy network. With the primitive knowledge of the previous environments in $\pi_f$, the robot agent is able to adapt to the new environment rapidly.

\subsection{Multi-objective Guided Rewards}

We propose a Multi-objective, Guided Rewards (MGR) system for the robotic task environment with sparse reward to improve training efficiency. In many robotic applications (e.g. pushing, peg-in-hole), a binary sparse reward is given to the robot agent depending on whether it achieved the desired goal. However, the sparse reward does not provide useful information to train the robot agent, so the exploration in the early stage is random and thus inefficient. 

The MGR is designed to encourage the agent to explore the state space, and also to guide the robot to the target with the estimated immediate rewards. The MGR system consists of three parts driven by three objectives: the final goal, the sequential objectives (e.g. initial objective, the secondary objective), and the prevention objective. The final goal is represented by the binary sparse reward judging whether the final desired goal is achieved. The sequential objectives are the objectives that guide the robotic behaviors in sequential phases to achieve the final goal. Finally, the prevention objective is to prevent any hindrance (e.g. obstacles, traps) during the task process. The general MGR is formulated as:
\begin{equation}
\begin{split}
     r  = & \alpha_1  G_f  + \sum_{i=2}^n \alpha_i O_i + \alpha_{n+1}  O_p
\end{split}
\end{equation}
where $G_f$ is the final goal, $N$ is the number of sequential objectives, $O_i$ is the sequential objectives, $O_p$ is the prevention objective, and $\alpha_i$ is the constant scale factor.  

Taking the robot pushing task as an example, the sequential objectives are decreasing the distance between the robot end-effector and the object $d_{oe}$ the distance between the object and target goal $d_{og}$. We also consider a new scenario for the pushing and sliding where there are obstacles on the table. So the prevention objective is also driven by moving away from the obstacles, increasing the distance between the robot end-effector and the obstacle $d_{es}$. The MGR for robot pushing task is formulated as:

\begin{equation}
\begin{split}
     r (d_{og}, d_{oe}, d_{es}) = & \alpha_1  (- \left \| d_{og}> \eta  \right \| )  +\alpha_2  (-d_{oe})   +
    \alpha_3  (-d_{og}) \\ & + ( \left \| d_{es}< \mu  \right \| ) (\log d_{es}-\log \mu)
\end{split}
\end{equation}
where $d_{og}$ is the distance between the object and the target goal, $d_{oe}$ is the distance between the object and the robot end-effector, $d_{es}$ is the distance between the robot end-effector and the obstacle, $\alpha_1, \alpha_2$ and $\alpha_3$ are weights for multiple objectives, $\eta$ is the distance threshold to measure whether a goal is achieved, and  $\mu$ is the distance threshold to measure whether an obstacle is too close.

\section{Experiment and Discussion}
\begin{center}
\begin{figure}[thpb]
    \centering
    \includegraphics[trim={0cm 0cm 0cm 0cm},clip,width=0.5\textwidth]{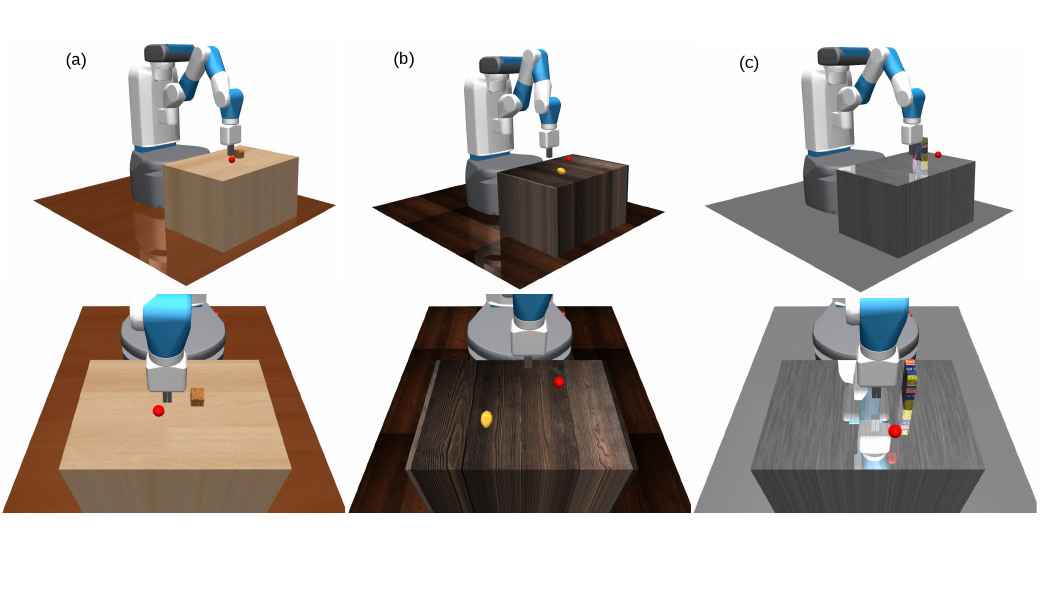}
    \caption{
    Robot pushing environments with different surfaces and different object shapes (a) Bread (b) Lemon  (c) Cereal Box
    }
    \label{envs}
\end{figure}
\end{center}

\subsection{Experiment Setup}
OpenAI gym \cite{brockman2016openai} simulation environment was used to test the proposed method. We implemented the tests with \textit{FetchPush} and \textit{FetchSlide} environment with the \textit{MuJoCo} \cite{todorov2012mujoco} physics engine. For each task, we also customized the environment settings with different surfaces, object shapes and obstacles to test the task generalization and environment adaptation, as shown in Figure~\ref{envs}. DDPG-HER algorithm was implemented and benchmarked based on the OpenAI stable-baselines implementation \cite{henderson2018deep}. A Multi-Layer Perception (MLP)-based policy network was implemented as the policy network in this work for the  model.

The finite episode length was set as 50 steps. In an episode, the robot agent receives a reward of $-1$ in each step if it did not achieve the desired goal, otherwise, it receives a reward of $0$. The Multi-objective, Guided Reward (MGR) function was implemented with the weights $\alpha_1 = 0.3$, $\alpha_2 = 0.35$ and $\alpha_3 =0.35$. The distances among the object, the target goal, and the robot end-effector were extracted from the simulation environment. In real-world, the distance information is usually measured with noise, so random noise was added to the distances $d_{og}$, $d_{oe}$ and $d_{es}$ during the tests. In this paper, we used robot pushing and sliding tasks as examples to test our algorithm, in a variety of different environment settings. 

\subsection{Results}

\begin{figure}[!thbp]
\centering
  \begin{subfigure}[b]{0.49\linewidth}
    \includegraphics[width=\textwidth]{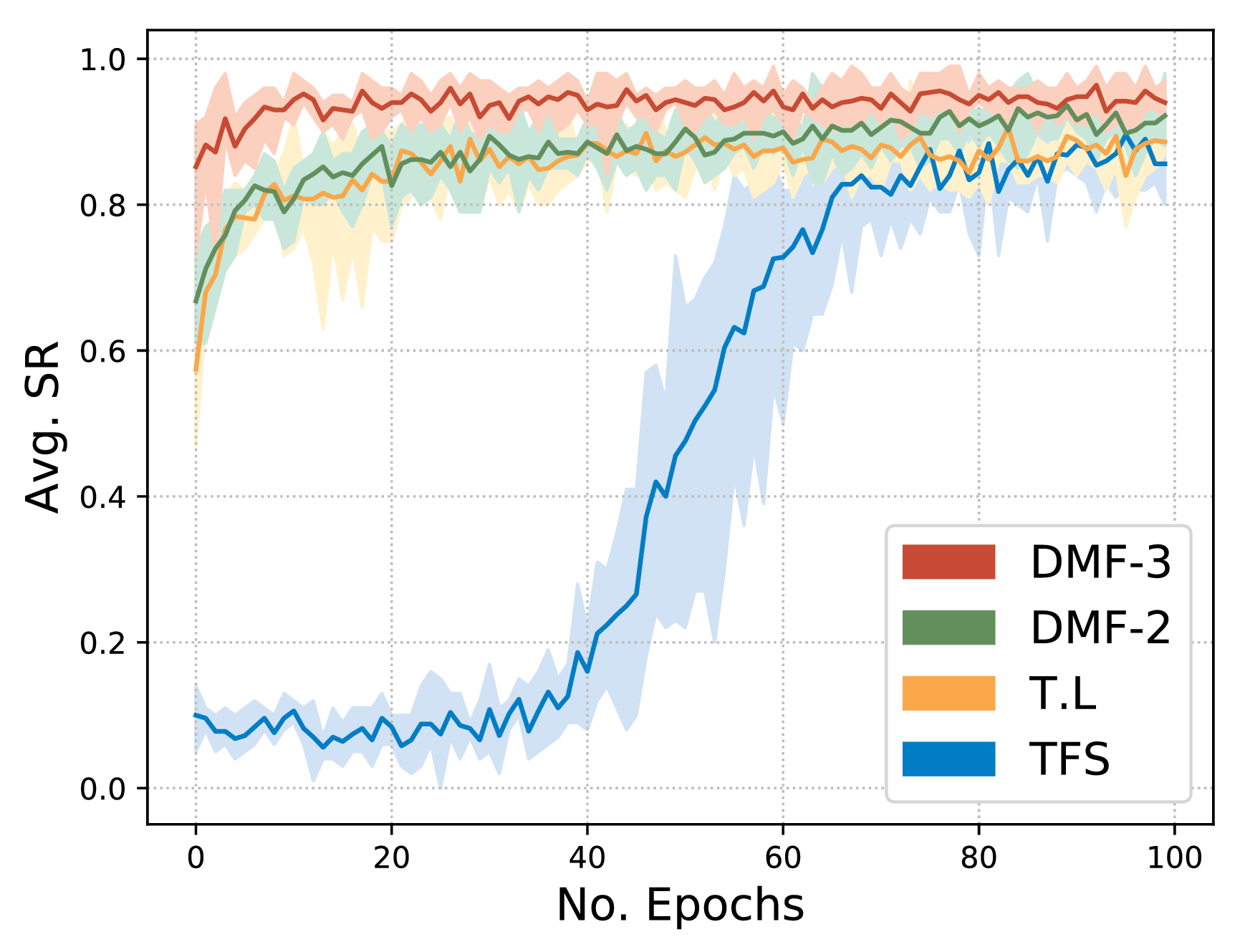}
    \caption{Average success rates}
    \label{f1}
  \end{subfigure}
  \begin{subfigure}[b]{0.49\linewidth}
    \includegraphics[width=\textwidth]{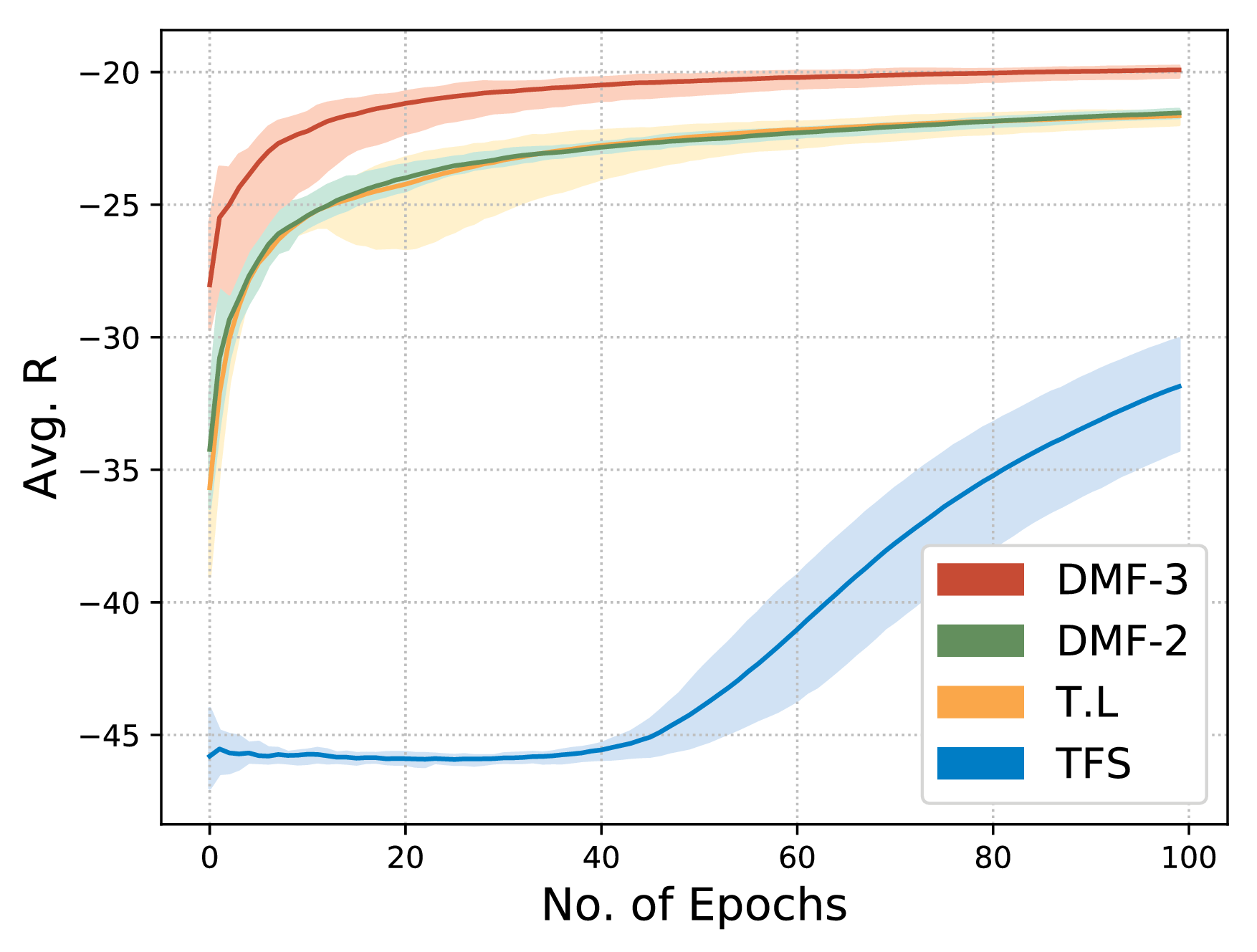}
    \caption{Average returns}
    \label{f2}
  \end{subfigure}
  \caption{Model Fusion for task generalization in robotic pushing task with different environment dynamics}
  \label{friction_modelfusion}
\end{figure}

\begin{table*}[th]
\centering
\scriptsize
\caption{Success rate comparison of different methods}
\begin{tabular}{c c c c c c c c c c c c c c c c c c}
\toprule
& & \multicolumn{4}{c}{ \textbf{DDPG-HER} } & \multicolumn{4}{c}{ \textbf{DDPG-HER + MGR} } & \multicolumn{4}{c}{ \textbf{DMF-2 + MGR} } & \multicolumn{4}{c}{ \textbf{DMF-3 + MGR} } \\ \midrule
& & 50 & 100 & 150 & 200 & 50 & 100 & 150 & 200 & 50 & 100 & 150 & 200 & 50 & 100 & 150 & 200 \\ \midrule
\multirow{3}{*}{ \textbf{Push} } & env-1 & 0.141 & 0.458 & 0.597 & 0.667 & 0.608 & 0.783 & 0.844 & 0.875 & 0.816 & 0.887 & 0.913 &  0.926 & 0.951 & 0.962 & 0.964 & 0.966\\
& env-2 & 0.175 & 0.518 & 0.648 & 0.718 & 0.445 & 0.657 & 0.733 & 0.773 & 0.828 & 0.859 & 0.875 & 0.885 & 0.868 & 0.892 & 0.9 & 0.904\\
& env-3 & 0.074 & 0.079 & 0.106 & 0.157 & 0.089 & 0.216 & 0.387 & 0.507 & 0.839 & 0.875 & 0.890 & 0.897 & 0.906 & 0.913 & 0.917 & 0.92\\ \midrule
\multirow{3}{*}{ \textbf{Sliding} } & env-1 & 0.206 & 0.342 & 0.411 & 0.451 & 0.342 & 0.513 & 0.597 & 0.645 & 0.424 & 0.576 & 0.647 & 0.679 & 0.662 & 0.726 & 0.755 & 0.774 \\
& env-2 & 0.098 & 0.158 & 0.224 & 0.291 & 0.183 & 0.296 & 0.376 & 0.425 & 0.302 & 0.509 & 0.606 & 0.659 & 0.618 & 0.725 & 0.766 & 0.79\\
& env-3 & 0.089 & 0.13 & 0.147 & 0.159 & 0.323 & 0.512 & 0.588 & 0.632 & 0.472 & 0.612 & 0.672 & 0.708 & 0.677 & 0.738 & 0.764 & 0.766 \\ \bottomrule
\end{tabular}\label{table:res}
\end{table*}

We first evaluated the Deep Model Fusion (DMF) and Multi-objective, Guided Reward (MGR) methods independently, before proceeding to evaluate the overall proposed method with both DMF and MGR. The results obtained in the tests show that the proposed method significantly improves the results in terms of learning speed and task success rate when adapting to the changed environment.   

\begin{figure}[!thbp]
\centering
  \begin{subfigure}[b]{0.49\linewidth}
    \includegraphics[width=\textwidth]{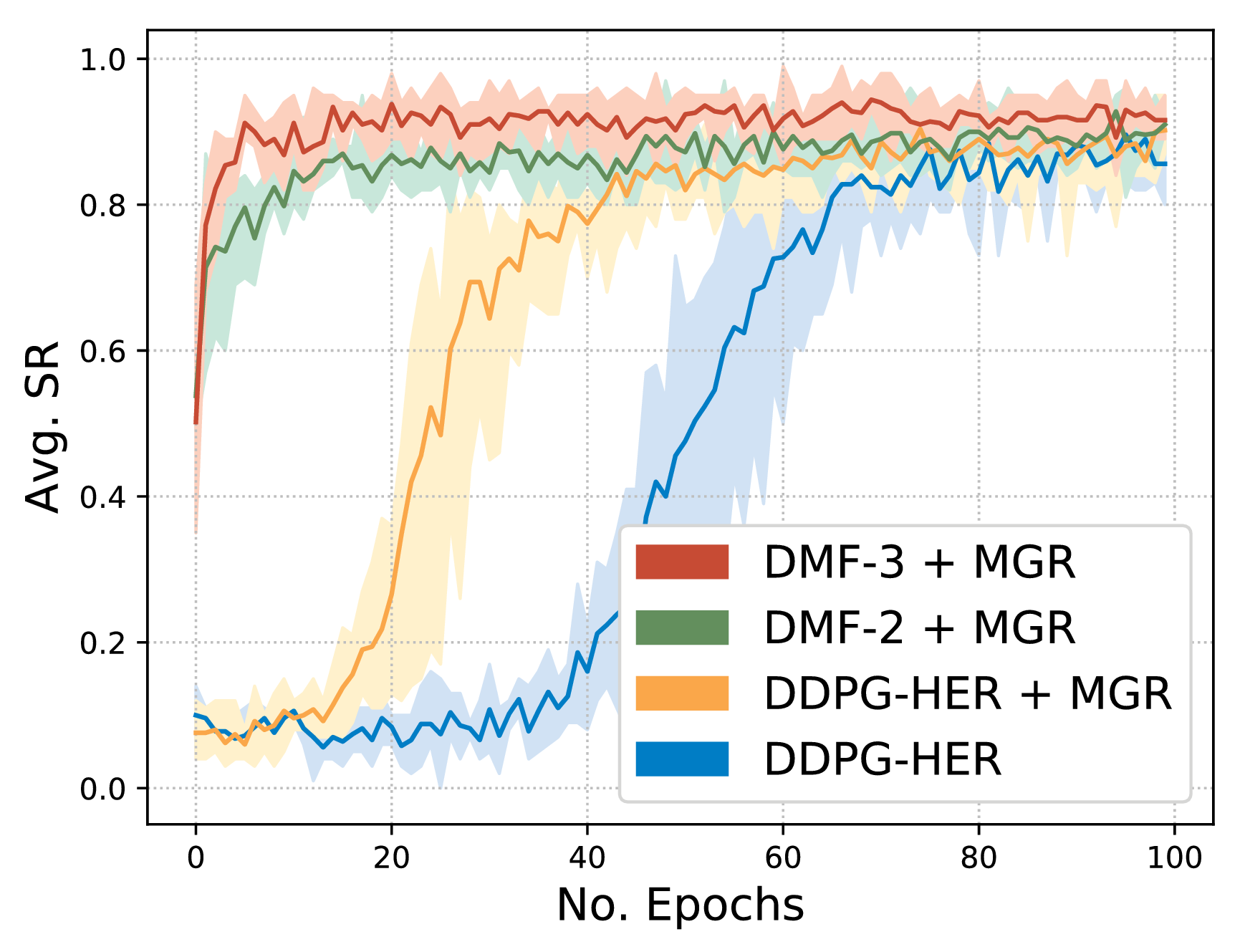}
    \caption{Average success rates}
    \label{f1}
  \end{subfigure}
  \begin{subfigure}[b]{0.49\linewidth}
    \includegraphics[width=\textwidth]{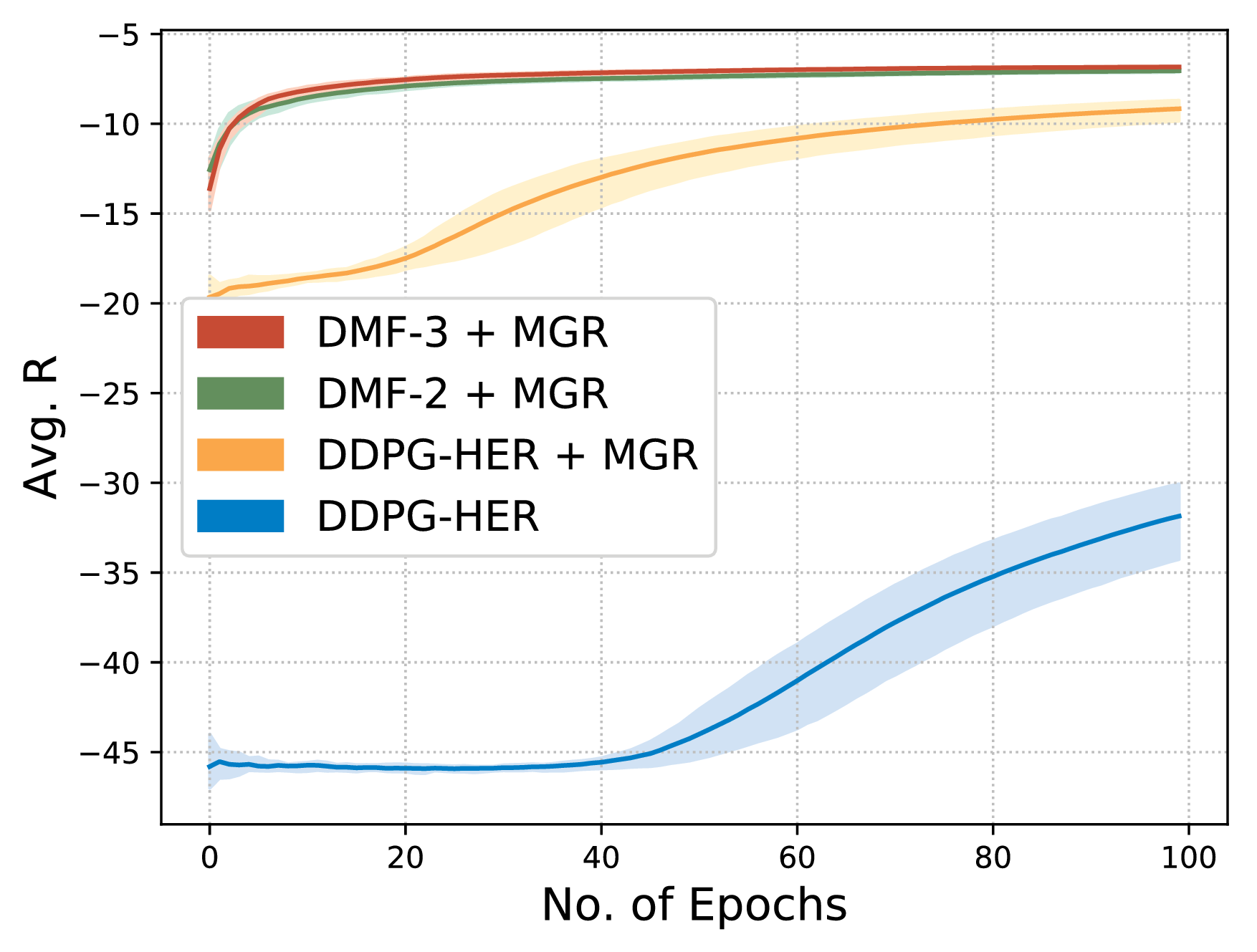}
    \caption{Average returns}
    \label{f2}
  \end{subfigure}
  \caption{Result comparison of MGR and DMF+MGR methods in robotic pushing application}
  \label{friction_mgr}
\end{figure}

The proposed method was evaluated in different environments (e.g. with different surfaces, different geometric shapes) for each task. We implemented the proposed DMF method that combined two models (labeled as DMF-2) and three models (labeled as DMF-3) and compared them with transfer learning (labeled as T.L), as well as training from scratch (labeled as TFS). Figure ~\ref{friction_modelfusion} shows the results in terms of average success rates and episodic returns in the robot pushing application. Compared to training from scratch and transfer learning, the robot agent learned faster with our method. The robot agent with primitive knowledge from our method demonstrated the good capability of task generalization and environment adaptation. The MGR was also evaluated with different environment settings. Figure ~\ref{friction_mgr} shows the results comparison of MGR and DMF+MGR methods in robot pushing application. Compared to baseline algorithms, the agent learned faster with the proposed MGR system.

Additionally, we further evaluated the overall proposed method with both DMF and MGR. Table \ref{table:res} shows the comparison of the results with different other methods at different training stages. We compared the baseline method DDPG-HER, DDPG-HER with MGR, as well as DDPG-HER with MGP and DMF (labeled as \textit{DMF-2 + MGR} for two-model fusion and \textit{DMF-3 + MGR} for three-model fusion). The success rates of the robotic tasks with different methods were compared under three different environment settings (labeled as env-1, env-2, and env-3). The success rates at different training stages (episodes = 50, 100, 150, 200) were also compared in the table. 

As shown in Table \ref{table:res}, the proposed method demonstrated its effectiveness among different applications in various training stages. The success rate of the proposed method was consistently higher than other methods, in different environments and different training stages.

\subsection{Discussion}

In the robotic task learning problem, the performance of the learned model usually drops when there are significant changes in the environment setting, though the learned policy is also able to adapt to certain change. It is also noticed that the transfer learning converges faster, compared to training from scratch. The proposed method that combines the knowledge from different models outperforms the other methods by both the learning speed and the success rate. 

Additionally, as shown in Figure \ref{friction_modelfusion}, DMF-3 and DMF-2 had very similar success rates and episodic returns, though DMF-3 converged faster than DMF-2. Both DMF-3 and DMF-2 significantly performed better than the baseline method. Although the other methods achieved a good success rate in some cases, their average episodic returns were still much lower. And as observed in the tests, the agent trained with the baseline method took a longer time to complete the task in one episode.

Overall, the proposed method outperformed the baseline algorithms. The results show that our method is able to generalize the learned robotic tasks efficiently by combining the knowledge in the previously trained models. The MGR system also helps to convert the sparse rewards to dense rewards with multiple objectives, where each objective could be interpreted intuitively with real-world correspondences. This feature makes the overall system explainable and robust. The proposed DMF and MGR could also be used as flavors on top of other RL algorithms to improve the performance. 

\section{Conclusion}

In this paper, we propose a novel Deep Model Fusion (DMF) method with Multi-objective Guided Reward (MGR) system for generalizing robotic task learning and environment adaptation. The proposed method improves the training efficiency of adapting the previously trained model to a new environment by combining knowledge from those models. Our method also improves the performance of task learning in terms of task success rate and average episodic return. The effectiveness of the proposed method has been validated by extensive studies in different environments settings.

\section*{Acknowledgment}

This research is supported by the Agency for Science, Technology and Research (A*STAR), Singapore, under its AME Programmatic Funding Scheme (Project \#A18A2b0046)

\bibliography{robio2019}

\end{document}